%% file: root.tex
\documentclass[letterpaper, 10 pt, conference]{ieeeconf}  % Comment this line out if you need a4paper

\IEEEoverridecommandlockouts                              % This command is only needed if 
\usepackage{amsmath} % assumes amsmath package installed
\usepackage{amssymb}  % assumes amsmath package installed
\usepackage{url}
\usepackage{caption}
\usepackage{subcaption}
\usepackage{cite}
\usepackage{siunitx}
\usepackage{pifont}% http://ctan.org/pkg/pifont

\usepackage{booktabs}
\usepackage{tabularx}
\usepackage{multirow}
\usepackage{xspace}
\usepackage{mathtools}
\usepackage{tikz}
\usepackage{pgfplots}
\pgfplotsset{compat=1.16}
\usepackage{adjustbox}
\usepackage{graphicx}
\usepackage[nogroupskip,acronyms,nopostdot,style=super,nonumberlist,toc]{glossaries}
\usepackage{cancel}
\usepackage{diagbox}
\usepackage{lipsum}
\usepackage{algorithm}
\usepackage{algpseudocode}

\newacronym{ai}{AI}{artificial intelligence}%
\newacronym{drl}{DRL}{deep reinforcement learning}
\newacronym{dl}{DL}{deep learning}
\newacronym{ml}{ML}{machine learning}
\newacronym{rl}{RL}{reinforcement learning}
\newacronym{ad}{AD}{autonomous driving}
\newacronym{av}{AV}{autonomous vehicle}
\newacronym{dnn}{DNN}{deep neural network}
\newacronym{ann}{ANN}{artificial neural network}
\newacronym{nn}{NN}{neural network}
\newacronym{dqn}{DQN}{deep Q-network}
\newacronym{cnn}{CNN}{convolutional neural network}
\newacronym{rnn}{RNN}{recurrent neural network}
\newacronym{rdqn}{RDQN}{recurrent deep Q-network}
\newacronym{ddqn}{DDQN}{double deep Q-network}
\newacronym{marl}{MARL}{multi-agent reinforcement learning}
\newacronym{dmarl}{DMARL}{deep multi-agent reinforcement learning}
\newacronym{mdp}{MDP}{Markov decision process}
\newacronym{mlp}{MLP}{multilayer perceptron}
\newacronym{mpc}{MPC}{model predictive control}
\newacronym{its}{ITS}{intelligent transportation systems}
\newacronym{ttc}{TTC}{time-to-collision}
\newacronym{ddpg}{DDPG}{deep deterministic policy gradient}
\newacronym{vae}{VAE}{variational auto-encoder}
\newacronym{mas}{MAS}{multi-agent system}
\newacronym{mal}{MAL}{multi-agent learning}
\newacronym{per}{PER}{prioritized experience replay}
\newacronym{a2c}{A2C}{advantage actor critic}
\newacronym{sg}{SG}{stochastic game}
\newacronym{mg}{MG}{Markov game}
\newacronym{pomdp}{POMDP}{partially observable Markov decision process}
\newacronym{pomg}{POMG}{partially observable Markov game}
\newacronym{dpomdp}{dec-POMDP}{decentralized partially observable Markov decision process}
\newacronym{nrmse}{NRMSE}{normalized root-mean-square error}
\newacronym{ppo}{PPO}{proximal policy optimization}
\newacronym{gae}{GAE}{generalized advantage estimate}
\newacronym{rpl}{RPL}{residual policy learning}
\newacronym{apf}{APF}{articial potential field}
\newacronym{lstm}{LSTM}{long short-term memory}
\newacronym{ftg}{FTG}{follow-the-gap}
\newacronym{il}{IL}{imitation learning}
\newacronym{rrt}{RRT}{rapidly-exploring random tree}
\newacronym{torcs}{TORCS}{The Open Racing Car Simulator}
\newacronym{cbf}{CBF}{control barrier function}
\newacronym{fov}{FOV}{field-of-view}
\newacronym{de}{DE}{disparity extender}

\newcolumntype{P}[1]{>{\raggedleft\arraybackslash}p{#1}}

\title{\LARGE \bf
RaceMOP: Mapless Online Path Planning for Multi-Agent\\Autonomous Racing using Residual Policy Learning
}

\author{Raphael Trumpp$^{1,2}$, Ehsan Javanmardi$^{2}$, Jin Nakazato$^{2}$, Manabu Tsukada$^{2}$, and Marco Caccamo$^{1}$% <-this % stops a space
\thanks{$^{1}$ TUM School of Engineering and Design, Technical University of Munich, Germany.}%
\thanks{$^{2}$ Graduate School of Information Science and Technology, The University of Tokyo, Japan.}
\thanks{Raphael Trumpp was a JSPS International Research Fellow. Marco Caccamo was supported by an Alexander von Humboldt Professorship endowed by the German Federal Ministry of Education and Research. This research was supported by JST ASPIRE Grant Number JPMJAP2325, Japan.}
}%

\begin{document}

\maketitle
\thispagestyle{empty}
\pagestyle{empty}

%%%%%%%%%%%%%%%%%%%%%%%%%%%%%%%%%%%%%%%%%%%%%%%%%%%%%%%%%%%%%%%%%%%%%%%%%%%%%%%%
\begin{abstract}
The interactive decision-making in multi-agent autonomous racing offers insights valuable beyond the domain of self-driving cars.
Mapless online path planning is particularly of practical appeal but poses a challenge for safely overtaking opponents due to the limited planning horizon.
To address this, we introduce RaceMOP, a novel method for mapless online path planning designed for multi-agent racing of F1TENTH cars.
Unlike classical planners that rely on predefined racing lines, RaceMOP operates without a map, utilizing only local observations to execute high-speed overtaking maneuvers.
Our approach combines an artificial potential field method as a base policy with residual policy learning to enable long-horizon planning. We advance the field by introducing a novel approach for policy fusion with the residual policy directly in probability space.
Extensive experiments on twelve simulated racetracks validate that RaceMOP is capable of long-horizon decision-making with robust collision avoidance during overtaking maneuvers.
RaceMOP demonstrates superior handling over existing mapless planners and generalizes to unknown racetracks, affirming its potential for broader applications in robotics.
Our code is available at \url{http://github.com/raphajaner/racemop}.
\end{abstract}

\input{source/introduction}

\input{source/related_work}
\input{source/background}
\input{source/methodology}
\input{source/results}

\input{source/conclusion}

\bibliographystyle{IEEEtran}
\bibliography{references}

\end{document}

%% file: source/introduction.tex
\section{Introduction}
Autonomous racing plays an essential role not only in creating self-driving cars \cite{wischnewski2022indy, betz2022autonomous} but serves as an ideal testbed for novel planning strategies for safe autonomy \cite{suresh2022threading}.
Especially racing with multiple agents is interesting for decision-making methods due to its nature as a non-cooperative \gls*{mas}.
So far, the challenging dynamic environment of racing has resulted in adopting path planners that predominantly rely on racing lines optimized offline using existing map data.
For example, the Indy Autonomous Challenge 2019 winning team employed an optimal racing line combined with a graph-based module for overtaking \cite{stahl2019multilayer}. 

We encourage extending the focus to mapless online path planning using only local observations from the vehicle's onboard sensors but no map data.
First, a mapless planner is more flexible and easily deployed to new environments, as the environment must not be mapped beforehand. 
Second, as only local observations are used, it is robust to local changes in the environment as no localization is required.
Results for F1TENTH competitions, a racing series for 1/10th scale autonomous RC cars \cite{okelly2020f1tenth}, support this claim. In these real-world races, mapless planners often secured victories due to their robustness \cite{otterness2019disparity}.
The same arguments apply to self-driving cars that must navigate unknown environments without map data, e.g., a rural road in a forest.
Additionally, mapless planning extends the scope of autonomous racing, serving as a testbed for general robotic path planning, e.g., mobile robots typically act in non-structured mapless environments.

After achieving superhuman performance in games \cite{mnih2013playing}, recent works, e.g., real-world drone racing \cite{kaufmann2023champion}, have shown that \gls{drl} can outperform humans also in challenging real-world decision-making problems.

\begin{figure}[t]
\vspace{2mm}
\centering
\includegraphics[width=0.42\textwidth]{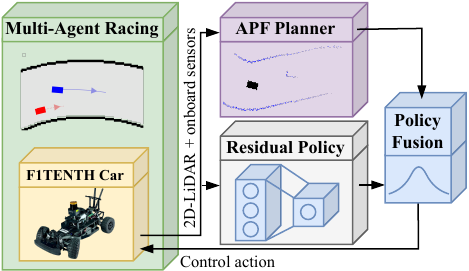}
\caption{Our method, named RaceMOP, is a novel mapless online path planner for multi-agent racing that uses only local observations. This method fuses an APF planner with a learned residual policy for simulated F1TENTH cars.}
\label{fig:cover_figure}
\end{figure}

This progress motivates us to propose {RaceMOP}, a \underline{M}apless \underline{O}nline \underline{P}ath planning method for autonomous multi-agent racing with F1TENTH cars \cite{okelly2020f1tenth}. As shown in Fig. \ref{fig:cover_figure}, RaceMOP uses \textit{only} 2D-LiDAR and onboard sensors but \textit{no} map data for planning overtaking maneuvers. 
Our method combines an \gls*{apf} planner, designed explicitly for multi-agent racing, with \gls*{rpl} to improve the vehicle's control action.
It has been demonstrated that \glspl*{apf} are effective methods to avoid collisions for self-driving cars \cite{lin2022local} while \gls*{rpl} is used successfully to improve driving policies in autonomous racing \cite{zhang2022residual, trumpp2023residual, evans2021learning}. 
\Gls*{rpl} trains a residual policy, parameterized by a \gls*{dnn}, with \gls*{drl} to amend the output of the base policy. This approach helps to solve decision-making tasks for which \gls*{drl} is data-inefficient or intractable while good but imperfect classical methods are available \cite{silver2018residual}.
RaceMOP's residual policy's \gls*{dnn} takes stacked past observations as input, allowing long-horizon decision-making when interacting with opponents.
This reasoning capability yields an advantage over classical path planners, which rely on precise localization and accurate predictions of future opponent trajectories and may otherwise fail to plan robust overtaking maneuvers.

Our main contributions can be summarized as follows:
\begin{itemize}
    \item We introduce RaceMOP, a mapless online path planning method using \gls*{rpl}, whic outperforms classical planners in simulated multi-agent autonomous racing.
    \item We present a novel method for policy fusion directly in probability space, enabling unbiased bounding of action spaces essential for effective \gls*{rpl}.    
    \item RaceMOP's generalization capabilities are validated through evaluation on twelve racetracks, including four previously unseen tracks. We discuss the learned driving behavior in a detailed scenario analysis.
    \item Our code and supplementary videos are available at {\small\url{http://github.com/raphajaner/racemop}}.
\end{itemize}

%% file: source/related_work.tex
\section{Related Work}
Multi-vehicle scenarios and overtaking, e.g., on highways, are well-studied in the literature for regular self-driving cars \cite{lin2022local}.
However, it poses a different challenge to autonomous racing due to the structured environment, e.g., traffic lanes and rules \cite{suresh2022threading}. For multi-agent autonomous racing, the main challenge is overtaking the non-cooperative opponents safely.

\subsection{Map-based Methods}
Access to map data allows tracking an offline-optimized racing line with an additional local planner for overtaking.
Searching an offline-generated multi-layer graph model to find feasible trajectories is effective in real-world racing \cite{stahl2019multilayer}. 
Raji et al. \cite{raji2022motion} use a global path as the main reference alongside a local Fren\'{e}t planner for overtaking in ovals.
A model predictive control method for F1TENTH cars capable of overtaking is presented by Li et al. \cite{li2022real}.

\subsection{Mapless Methods}
Closest to our work is Zhang et al. \cite{zhang2022residual}, who discuss an \gls*{apf} planner with residual control for \textit{single}-agent racing.
Their framework does not require object detection of walls or other vehicles since they simply define each LiDAR point as an obstacle in the \gls*{apf}.
Our approach extends this basic concept in various aspects, allowing us to focus on challenging multi-agent scenarios.
We discuss further mapless methods for \textit{multi}-agent racing in the following.

\subsubsection{Classical Mapless Methods}
A popular choice in F1TENTH racing is the \gls*{ftg} controller introduced by Sezer et al. \cite{sezer2012novel}. This reactive method steers the car toward a gap within a LiDAR reading. Similarly, the disparity extender \cite{otterness2019disparity} is based on detecting gaps but modifies the LiDAR's distance values to account for the vehicle's size at disparities. Another popular choice is the \gls*{rrt} method and its derivatives \cite{karaman2010incremental}. These methods generate a tree of trajectory candidates with forward dynamics, checked for feasibility, and ranked according to their cost \cite{feraco2020local, ma2014fast}.
The tentacles method \cite{von2008driving} is an empirical approach that draws tentacles as a geometric shape in the ego vehicle's centered reference frame. Given an egocentric occupant grid, \cite{alia2015local} proposes clothoid tentacles that are checked for feasibility. The local motion planner for the 2005 DARPA challenge used by Thrun et al. \cite{thrun2006stanley} is based on a similar collision avoidance method with tentacles parallel to the base path.
Uusitalo and Johansson \cite{uusitalo2011reactive} present an \glspl{apf} method in TORCS with multiple opponents.

\subsubsection{Learning-based Mapless Methods}
Loiacono et al. \cite{loiacono2010learning} present a controller for overtaking maneuvers in TORCS learned by reinforcement learning. While Fuchs et al. \cite{fuchs2021super} show that \gls*{drl} is capable of achieving superhuman performance in Gran Turismo by developing a mapless \gls*{drl} controller, their model is limited to time-trial races without opponents.
Song et al. \cite{song2021autonomous} show that curriculum \gls{drl} is needed to learn overtaking maneuvers from LiDAR.
However, their agent observes the angle and curvature of the track's centerline, and their method comes at the cost of a hand-crafted training scheme with additional hyperparameters. 
Moving the focus to the F1TENTH cars, Zhang and Loidl \cite{zhang2023f1tenth} present an over-taking algorithm using behavior cloning and \gls*{rnn} but with access to the opponent's true state.
Evans et al. \cite{evans2021learning} introduce an \gls*{rpl} approach for collision avoidance of static objects.
A multi-team racing scenario is discussed by Werner et al. \cite{werner2023dynamic}.
Their simulator merges an analytic model with a data-driven dynamic model to be used in the real world with a hierarchical policy structure.
Schwarting et al. \cite{schwarting2021deep} learn a world model in latent space to imagine self-play that reduces sample generation in a multi-agent \gls*{drl} training scheme.
The work of Kalaria et al. \cite{kalaria2023towards} proposes curriculum learning to utilize a control barrier function that is gradually removed during training to not comprise the final performance. 

%% file: source/background.tex
\section{Technical Background}
The following section introduces the \gls*{apf} method for path planning and presents the \gls*{drl} and \gls*{rpl} theory, respectively.

\subsection{Artificial Potential Fields Planner}
\Gls*{apf} path planners are used for collision-free robot motion planning.
The path is planned by defining a potential field that consists of repulsive forces for obstacles to be avoided, while the goal is defined as an attractive well \cite{park2001obstacle}.
Typically, a quadratic potential for repulsive forces is chosen as
\begin{equation}
\label{eq:rep_field}
U_{\text{rep}}(x)= \begin{cases}\frac{1}{2} k_{\text{rep}}\left(\frac{1}{\rho}-\frac{1}{\rho_0}\right)^2 & \text { if } \rho \leq \rho_0 \\ 0 & \text { if } \rho>\rho_0\end{cases},
\end{equation}
where $\rho$ is the obstacle's minimal distance to the robot's 2D position $x$. A threshold $\rho_0$ limits the area of influence, and $k_{\text{rep}}$ is a gain factor \cite{khatib1985real}. The attractive well has the form 
\begin{equation}
\label{eq:att_field}
U_{\text{att}}(x)=\frac{1}{2} k_{\text{att}}\left|x-x_d\right|^2,
\end{equation}
with $k_{\text{att}}$ as attractive gain leading to the definition of the full potential field as $U = U_{\text{att}}(x) + \sum U_{\text{rep}}(x)$.
The path is then iteratively obtained by following the potential's gradient
\begin{equation}
\label{eq:step_apf}
    x' = x - \epsilon \frac{\nabla U}{||\nabla U||_2},
\end{equation}
with step size $\epsilon$ to the next position $x'$ minimizing $U$.

\subsection{Deep Reinforcement Learning}
The \gls*{mdp} for model-free \gls*{drl} is defined by the tuple $(\mathcal{S}, \mathcal{A}, \mathcal{T}, \mathcal{R}, \gamma)$. The stochastic action policy $\pi_{\theta}(a_t | s_t)$ is parameterized by $\theta$; no forward dynamics need to be explicitly defined.
This policy maps observed states $s_t \in \mathcal{S}$ to a probabilistic action space $\mathcal{P}(\mathcal{A})$ of (continuous) actions $a_t \in \mathcal{A}$.
The probability for transitioning from $s_t$ to $s_{t+1}$ is defined by $\mathcal{T}: \mathcal{S} \times \mathcal{A} \rightarrow \mathcal{P}(\mathcal{S})$, leading to a scalar reward value $r_{t+1}$ of the reward function $\mathcal{R}$.
The discount factor is defined as $\gamma$. 
The goal in \gls*{drl} is to find the parameters $\theta$ of the optimal policy $\pi_{\theta}^{*}(a_t | s_t)$ that maximizes the expected return $V_{\pi_{\theta}}(s_t) = \mathbb{E}_{\pi_{\theta}}\left[\sum_{k=0}^{\infty} \gamma^k r_{t+k+1} \mid s_t\right]$.  

\Gls*{ppo}~\cite{schulman2017proximal} is an \textit{on}-policy \gls*{drl} method used to learn a stochastic policy $\pi_{\theta}(a_t | s_t)$, where $\theta$ are the weights of a \gls*{dnn}.
The weights $\theta$ are updated with respect to the advantage function, which requires estimating the value function $V_{\phi}(s_t)$, parameterized by a separate set of weights $\phi$.

\subsection{Residual Policy Learning}
Silver et al. \cite{silver2018residual} introduce \gls*{rpl} as the combination of classical controllers with a residual policy learned by \gls*{drl}. 

For generality, we describe this approach by defining the action policy $\pi_{\theta}(a_t | s_t)$ as the result of an operation that fuses a deterministic \textit{base} policy $\mu_{\text{B}}(s_t): \mathcal{S} \to \mathcal{A}$ with a learned stochastic residual policy $\pi_{\text{R}, \theta}:  \mathcal{S} \to \mathcal{P}(\mathcal{A})$, parameterized by the weights $\theta$ of a \gls*{dnn}. 
Let the operator $\circledast: \mathcal{A} \times \mathcal{P}(\mathcal{A}) \rightarrow \mathcal{H}$ be defined such that the policy fusion is a mapping 
\begin{equation}
    h = \mu_{\text{B}}(s_t) \circledast \pi_{\text{R},\theta}(\cdot| s_t),
\end{equation}
to parameters $h \in \mathcal{H}$ of a parameter space $\mathcal{H}$.
We then define the action policy as a probability distribution $D(h, \Lambda)$ by
\begin{equation}
\label{eqn:residual_def}
    \pi_{\theta}(\cdot | s_t) := \mathcal{D}(h, \Lambda),
\end{equation}
with $h$ as the central parameter of the distribution plus possible constraints $\Lambda$.
Actions $a_t$ are sampled $a_t \sim  \mathcal{D}(h, \Lambda)$ and the weights $\theta$ of policy $\pi_{\theta}(a_t | s_t)$ are learned by \gls*{drl}.

This definition accommodates trivial expressions for $a_t$, including the sum $a_t = a_{\text{R}, t} |_{a_{\text{R}, t} \sim \pi_{\text{R}, \theta}(\cdot | s_t)} + \mu_{\text{B}}(s_t)$, as used in \cite{trumpp2023residual}, by defining $D$ as a Dirac delta distribution, but also captures advanced fusion operators such as the truncated Gaussian used in this work.
For effective learning, $D(h, \Lambda)$ should ensure that the action policy follows the base policy without bias when the residual part is off, i.e., $a_t \approx a_{\text{B}, t}$.

%% file: source/methodology.tex
\section{Methodology}
\label{sec:methodology}
This work introduces RaceMOP, a mapless online path planner for multi-agent autonomous racing in the F1TENTH gym simulator \cite{okelly2020f1tenth}, that consists of two modules (see Fig. \ref{fig:cover_figure}): 
 \begin{itemize}
    \item \textit{Base policy} $\mu_{\text{B}}(s_t)$: An \gls*{apf} planner with an action $a_{\text{B},t}$ designed for multi-agent racing; see Sec.~\ref{subsec:base_controller}.
    \item \textit{Residual policy} $\pi_{\text{R},\theta}(a_t | s_t)$: A stochastic action policy with learnable weights $\theta$ of a \gls*{dnn}. The novel policy fusion with the base planner is presented in Sec.~\ref{subseq:racemop}.
\end{itemize}

The methodology is developed under the following assumptions of a race environment as non-cooperative \gls*{mas}:
\begin{enumerate}
    \item An overtaking maneuver consists of only two agents.
    \item Only the ego vehicle is overtaking, i.e., no defensive or adversarial behavior to defend its position is needed.
    \item All opponent vehicles track a racing line in a non-reactive way, which aligns with the literature \cite{suresh2022threading, zhang2023f1tenth}.
\end{enumerate}

\subsection{Artificial Potential Fields Planner}
\label{subsec:base_controller}
\begin{figure*}[!ht]
\vspace{1.5mm}
    \begin{minipage}[t]{0.48\textwidth}
       \centering
        \includegraphics[height=2.4cm]{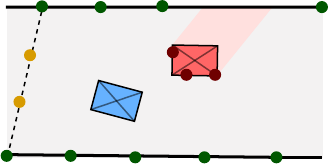}
        \caption{The ego vehicle (blue) perceives the environment by LiDAR points; green points show the wall (black line). When another vehicle (red) is present, its shape is approximately reflected by LiDAR points (red), but parts of the wall get occluded (red area). After filtering the points, the gap behind the ego vehicle is closed with artificial points (orange). }
        \label{fig:lidar}
    \end{minipage}%
    \hfill
    \begin{minipage}[t]{0.48\textwidth}
        \centering
        \includegraphics[height=2.4cm]{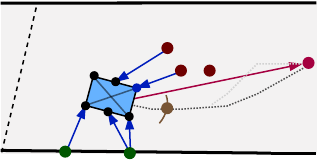}
        \caption{Only a subset of LiDAR points (blue) closest to the ego vehicle's edge points (black) are considered as repulsive forces (blue arrows) for the \gls*{apf}, while the goal point (purple) is attractive. The calculated path (light gray) is smoothed (dark gray) to avoid abrupt direction changes. The tracking point (brown) is found in a fixed lookahead distance.}
        \label{fig:apf}
    \end{minipage}%
\end{figure*}

The F1TENTH cars perceive their environment by a 2D-LiDAR in a fixed \gls*{fov} with $i$ points $p_{i} = (\alpha, d)_{i}$ at angle $\alpha$ and distance $d$.
Fig. \ref{fig:lidar} visualizes the LiDAR signal projected to the x-y-plane.
All LiDAR points can be defined directly as repulsive forces in an \gls*{apf} for collision avoidance \cite{zhang2022residual}, alleviating the need to detect the opponent's position.
This naive \gls*{apf} method fails for multi-agent racing due to the interaction with opponents and non-holonomic vehicles. 
We overcome these limitations as follows. 

\subsubsection{Down-sample Filtering}
\label{subsubsec:down_sampling}
We deploy a filtering that starts from the first LiDAR point at the minimum left angle.
The next point $p_{i+t}$ gets only added to the list $P=\{p_{0}, ..., p_{i}\}$ when $||p_{i+t}-p_{i}||_{2} > \epsilon_{f}$ with threshold $\epsilon_{f}$.
This filtering produces a smoothed x-y-representation and reduces computational resources in the following steps. 

\subsubsection{Closing the Gap}
It rarely occurs that the \gls*{apf}'s gradient leads to a path outside the walls.
This is due to the LiDAR's limited \gls*{fov} and occlusion in curves.
We prevent such paths by first removing all points in distance $d_f$ behind the car and then adding linearly interpolated artificial points between the first and last LiDAR points behind the car.

\subsubsection{Goal Point Tracking}
Inspired by the \gls{ftg} approach, we first identify all possible gaps. A gap is described by a disparity in the LiDAR signal exceeding a threshold $\epsilon_{d}$. Once all gaps are found, the candidate goal points are defined by a ray $p=(\max(d_{i},d_{i+1}), (\alpha_{i}+\alpha_{i+1})/2)$. We select the point the furthest away from the ego as the goal point $p_g$.

\subsubsection{Local Collision Avoidance}
Instead of modeling the ego size as a point mass or circle, we define six points of the vehicle's body to calculate the \gls*{apf}. For each of these points, we then consider only the repulsive forces from the closest point in the filtered set of LiDAR points as visualized in Fig.~\ref{fig:lidar}. 
Experiments have shown that this improves overtaking as the car's size is correctly reflected at its sides, which is not the case when approximating the car's size as a circle.

\subsubsection{Linear Potentials}
Instead of the quadratic potentials as introduced in (\ref{eq:att_field}) and (\ref{eq:rep_field}), we use a linear potential
\begin{equation}
\label{eq:rep_field_lin}
U_{\text{rep}}(x)= \begin{cases} k_{\text{rep}}\left(\frac{1}{\rho}-\frac{1}{\rho_0}\right) & \text { if } \rho \leq \rho_0 \\ 0 & \text { if } \rho>\rho_0\end{cases},
\end{equation}
while the attractive well has the form 
\begin{equation}
\label{eq:att_field_lin}
U_{\text{att}}(x)=k_{\text{att}}\left|x-x_d\right|.
\end{equation}
Our experiments show this choice helps because it keeps the attractive force constant. Otherwise, a goal point far away would lead to a stronger attractive force than a close one, which hinders local collision avoidance as the planned path typically tries to avoid the local obstacle with minimal detour in the direction of the goal. This leads to unsmooth paths that the non-holonomic F1TENTH cars cannot properly follow.

\subsubsection{Path Planning and Smoothing}
The planned path ${(x,y)_{i+i}, ..., (x,y)_{i+n_p}}$ is obtained by iteratively using (\ref{eq:step_apf}) to follow the \gls*{apf}'s gradient $n_p$ times while the environment is assumed to be static.
These raw paths are often not smooth. 
Additionally, \gls*{apf} can get stuck in local minima in symmetric passages \cite{park2001obstacle}, e.g., when there is only a narrow gap when overtaking, leading to a zick-zack path.
We apply the filtering method described in Sec. \ref{subsubsec:down_sampling} again and then further smooth the path by fitting a cubic spline.
The tracking point $p_{t}$ is subsequently determined from the cubic spline using the lookahead distance $l_{t}$ as shown in Fig. \ref{fig:apf}. 

\subsubsection{Control Command with Velocity Profile}
Given the tracking coordinate $p_{t}=(x,y)$, a pure pursuit  \cite{coulter1992implementation} controller is used to obtain the steering command $\delta_t$. Calculating the target velocity is challenging as it must combine the vehicle's physical properties with planning for safe overtaking. We propose to calculate the target velocity $v_t$ as a minimum
\begin{equation}
    v_t = \min \left(v_{1}(\mu_f, \delta_t), v_{2}(d_g)\right) %v_{2}(\delta_t),
\end{equation}
with $v_{1}(\mu_f, \delta_t)$ reflecting the physical friction limit \cite{evans2023safe} by
\begin{equation}
     v_{1}(\mu_{f}, \delta_t) = \sqrt{\frac{\mu_{f} l g}{\tan |\delta_t|}},
\end{equation}
given the friction coefficient $\mu_{f}$, the vehicle's wheelbase $l$, and the gravity constant $g$.
% A lookup table is used for $v_{2}(\delta_t)$ to decrease the velocity when large steering angles are calculated. 
The distance $d_{g}$ to the goal point $p_g$ defines $v_{2}(d_g)$ to model the interaction when overtaking. 
During overtaking, a goal point close to the ego vehicle indicates that there is no or only a narrow passage for overtaking. Therefore, $v_{2}(d_g)$ is proportionally decreased as it is assumed that no safe overtaking is possible.

\subsection{RaceMOP}\label{subseq:racemop}
RaceMOP uses the presented \gls*{apf} planner as base policy $\pi_{\text{B}}$ with corresponding action $a_{\text{B}, t}$.
We implement the residual policy $\pi_{\text{R}, \theta}$ as a \gls*{dnn} that learns the parameters of a yet-to-be-defined probability distribution, namely a state-dependent parameter $\mu_{R,\theta_{1}}:=f(s_t ; \theta_{1})$ and another parameter $\sigma_{R, \theta_{2}}:= \theta_{2}$, i.e., the learnable parameters are $\theta=(\theta_{1}, \theta_{2})$. 
The action policy's weights $\theta$ are learned by the \gls*{ppo} algorithm as later outlined in Sec. \ref{seq:learning_algo}.

\subsubsection{Policy Fusion}
\label{subsubsec:controller_conv}
In reference to the arbitrary $\mathcal{D}(h,\Lambda)$ in (\ref{eqn:residual_def}), we aggregate $\pi_{\text{R}}$ and $\mu_{\text{B}}$ in RaceMOP by means of a \textit{truncated} Gaussian distribution $\mathcal{N}(\mu, \sigma, c^{-}, c^{+})$ \cite{burkardt2014truncated} with 
\begin{align}
    \mu &= a_{\text{B}, t}  + \alpha \cdot \mu_{R,\theta_{1}}\label{eq:mu}\\
    \sigma &= \sigma_{R,\theta_{2}}.
\end{align}
The truncation interval $[c^{-}, c^{+}]$ ensures that sampled actions are bounded as $a_{t} \in [c^{-}, c^{+}]$. This distribution's mean $\mu$ is also its mode. Therefore, when $\mu_{R,\theta_{1}}=0$ and the mode is sampled, then $a_{t} = a_{\text{B}}$.
Using other distributions can lead to a biased mapping of the base action when fusing before the probability function or wrong gradients when clipping is used for bounded actions. 
Using the truncated Gaussian overcomes both limitations and correctly redistributes the probability mass for the truncation interval.

\subsubsection{Action And State Space}
RaceMOP's continues action $a_{t}=[v_t, \delta_t]$ consists of the target speed $v_t$ and steering angle $\delta_t$ as the input to the car's low-level controller. Bounded actions are sampled $ a_{t} \sim \mathcal{N}(\mu, \sigma, c^{-}, c^{+})$ with $c^{\pm}=[\pm1,\pm1]$ as required by the low-level controller.

Only \textit{local} information is used for the state
\begin{equation}
    s^{o}_{t} = 
    \begin{bmatrix*}
        L_t & v^{\text{long}}_{t} & v^{\text{lat}}_{t} & \dot v^{\text{long}}_{t}  & \dot\psi_{t} & \beta_{t} & \delta_t & a_{t-1}
    \end{bmatrix*}^\top,
\end{equation}
with the 2D-LiDAR $L_t\in\mathcal{R}^{1080}$ covering a 270$^{\circ}$ \gls*{fov}. The ego's dynamic state is represented by the longitudinal velocity $v^{\text{long}}_{t}$, lateral velocity $v^{\text{lat}}_{t}$, longitudinal acceleration $\dot v^{\text{long}}_{t}$, the yaw rate $\dot\psi$, the slip angle $\beta$, steering $\delta_t$, and the previous action $a_{t-1}$. 
Temporal information is embedded by stacking $n_{f}$ previous states where $n_s$ frames are skipped in between, forming the current state as $s_t = \{ s^{o}_{t-n_{f}(1+n_s)}, \dots, s^{o}_{t}\}$.

\subsubsection{Reward Design}
Our reward is independent of map data by using the ego's longitudinal velocity to calculate the traveled distance  $d^{\text{long}}_{t}=v^{\text{long}}_{t} \cdot \Delta t$ between timesteps of duration $\Delta t$ as the main optimization target:
\begin{align}
    r_{t+1} =&\, 0.1 \cdot d^{\text{long}}_{t} - 0.005 \cdot |a_{t} - a_{t-1}|  \nonumber \\
     & - 0.2 \cdot d_{L,t} \cdot \mathbf{1}_{d_{l}} + 0.5 \cdot \mathbf{1}_{o} - 5 \cdot \mathbf{1}_{c}.
     \label{eq:reward}
\end{align}
Large action changes between time steps are penalized, as well as small distances to obstacles with $\mathbf{1}_{d_{l}}=1$ when $d_{L,t} = \min L_{t} < 0.4$. Successful overtaking are encouraged by $\mathbf{1}_{d_{0}}=1$ and collisions are penalized $\mathbf{1}_{c}=1$.

\subsection{Learning Algorithm and Network Design}
\label{seq:learning_algo}
\begin{figure}[!t]
    \centering
    \vspace{1.5mm}
    \includegraphics[width=0.43\textwidth]{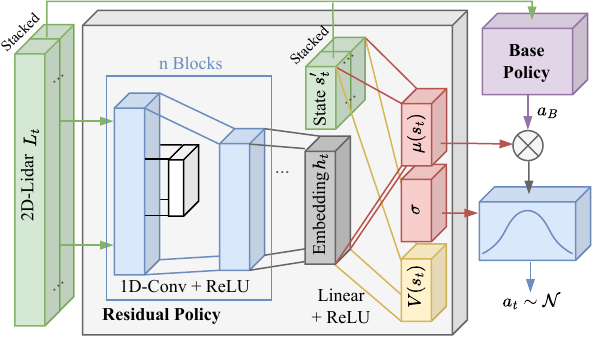}
    \caption{RaceMOP's architecture combines a base policy with a residual policy to learn the parameters of a probability distribution $\mathcal{N}$ from only local observations $s_t=\{L_t, s'_{t}\}$ that contains a history of $n_f$ frames.}
    \label{fig:network}
\end{figure}

RaceMOP is based on a \gls*{ppo} agent \cite{schulman2017proximal} with action policy $\pi_{\theta}$, value network $V_{\phi}$, and learnable parameters $\{\phi, \theta\}$. As shown in Fig. \ref{fig:network}, the LiDAR signal $L_t$ is encoded by a shared \gls*{dnn} consisting of $n=5$ blocks of 1D-\gls*{cnn} layers (\#-filters, size, stride) with $\{(64, 6, 4), (128, 3, 2), (256, 3, 2), (256, 3, 2), (256, 3, 2))\}$, respectively, and ReLU activation. The state is built by stacking $n_f=6$ frames to the current one, skipping $n_s=2$ frames in between. The temporal structure in $L_t$ is reflected by separating the different timesteps into separate input channels. The embedding $h_t$ is obtained by a linear projection layer with 64 neurons after the 1D-\gls*{cnn} layers and concatenated to the remaining flattened states $s'_t$. Two separate network heads for the policy and value network are used, consisting of linear layers with ReLU activation and $\{256,2\}$ and $\{256,1\}$ units, respectively. This yields 828,293 trainable parameters overall. The policy's output is then fused with the base policy by a truncated Gaussian as discussed in Sec.~\ref{subsubsec:controller_conv}.

%% file: source/results.tex
\section{Results}
We benchmark RaceMOP in Sec. \ref{subsec:results_1} and present additional results for generalization in Sec. \ref{subsec:results_2}.
RaceMOP's learned driving behavior is analyzed for various racing situations, including visualizations, in Sec. \ref{subsec:results_3}.

\subsection{Setup Experiments}
\subsubsection{Simulator Settings}
We employ a custom F1TENTH gym simulator \cite{okelly2020f1tenth}.
The used maps are replicas of famous real-world racetracks, appropriately downscaled for F1TENTH cars \cite{betz2022autonomous}.
These maps feature challenging design elements like chicanes and hairpin curves.
All opponents use a pure pursuit controller to follow an offline-optimized racing line with velocity profile $\{v_{w_0}, v_{w_1}, ...\} \cdot k_{v}$.
The velocity gain $k_{v}$ limits the opponent's velocity.
We define 30 start positions for each racetrack.
Episodes are truncated after two laps.
See Sec. \ref{sec:methodology} for further assumptions.
The simulation is based on a single-track model with tire slip and a maximum velocity of $\SI{8.0}{\meter \per \second}$.
We use a friction coefficient of $\mu_{f}=0.8$ to ensure there is slip when driving too fast in curves.
The observation and control frequencies are fixed to $\SI{50}{\hertz}$, similar to real-world F1TENTH cars.
For numerical stability, the simulation runs at $\SI{100}{\hertz}$.
The LiDAR data contains Gaussian noise.

\begin{table}[!t]
\vspace{1.5mm}
\centering
\begin{tabular}{l | l r l r }
\toprule
    \multirow{7}{*}{\rotatebox[origin=c]{90}{\textbf{PPO}}} 
    & Total steps & 30e\textsuperscript{6} & Discount $\gamma$ & 0.99 \\ 
    & Learn. rate (LR) &  1e\textsuperscript{-4} & GAE $\lambda$ & 0.95 \\
    & LR schedule & $\cos$ &  Value coef. & 0.5 \\
    & Traj. length & 2048 &  Max. grad. norm & 1.0 \\
    & Clip $\epsilon$ & 0.1 & Update epochs & 7 \\
    & Batch size & 512 & RPO\textsubscript{$\alpha$} \cite{rahman2022robust} & 0.05 \\
    & Initialization $\sigma_{R,\theta_{2}}$ & -0.7 & $\alpha$ (see (\ref{eq:mu}) & [0.5, 0.5]\\
\midrule
    \multirow{5}{*}{\rotatebox[origin=c]{90}{\textbf{APF}}} 
    & Attractive coeff. $k_{\text{att}}$  & 1000 & $\rho_0$ & 8.0 \\ 
    & Repulsive coeff. $k_{\text{rep}}$ & 25 & Step size $\epsilon$ & 0.1 \\
    & Lookahead target $l_t$ & 1.0 &  Gap distance $\epsilon_d $ & 1.0 \\
    & Filtering $\epsilon_f $ & 0.1 & Filtering $d_f$ & -4.0 \\
    &  Iterations $n_p$ & 20 \\
\bottomrule
\end{tabular}
\caption{Hyperparameters used in RaceMOP's modules.}
\label{table:hyperparameters}
\end{table}

\subsubsection{Training Procedure}
\label{sec:training-procedure}
RaceMOP is trained for $30e^{6}$ simulation steps.
Experiments have shown that high parallelization with 256 environments is beneficial.
Training is randomized by racing on eight different racetracks.
Additionally, $k_{v} \sim \{0.8, 0.75, 0.7\}$ is sampled uniformly every episode for all opponents in the same environment, and new random starting positions are chosen.
Thus, the ego vehicle must learn robust overtaking strategies for various scenarios, making the task more challenging.
Overtaken opponents are reset to a close position in front of the ego.
Observations and rewards are normalized.
We use a median filter to remove noise in the LiDAR data before being fed into the \gls*{dnn}. Other important hyperparameters of RaceMOP and for implementing the \gls*{apf} base policy can be found in Table~\ref{table:hyperparameters}. Full training takes approx. \SI{3.5}{h} on a server with AMD TR PRO 5975WX CPU and NVIDIA GeForce RTX 4090 GPU.

\subsubsection{Evaluation Details}
To analyze the RaceMOP's generalization capabilities, we evaluate our model for twelve racetracks, i.e., the eight training racetracks and four additional ones for testing. Reported results are the mean and standard deviation from five different training seeds. Each independent run is evaluated for thirty episodes with varying start positions and averaged as medians for lap times, while the other metrics are means. Nine opponents are distributed along the racetracks with a gain fixed to $k_{v}=0.75$ for the evaluation. 
We define the following three metrics:
\begin{itemize}
    \item $I_{T} [\downarrow]$: Lap time in $\si{\second}$ as duration of the running start lap.
    \item $I_{C} [\downarrow]$: The crash rate of the ego when attempting to overtake as $\frac{\#_{\text{crash}}}{\#_\text{success} + \#_\text{crash}} |_{\text{overtaking}}$ in \%.
    \item $I_{E} [\downarrow]$: Environment crashes of the ego while not overtaking over the total driven distance in kilometer.
\end{itemize}

\subsection{RaceMOP Benchmarking}
\label{subsec:results_1}

\begin{table*}[!t]
\centering
\vspace{1.5mm}
\begin{tabular}{c l | S[table-format=2.2] S[table-format=2.2] S[table-format=2.2] | S[table-format=2.2] S[table-format=2.2] S[table-format=2.2] | S[table-format=2.2] S[table-format=2.2] S[table-format=2.2] }
\toprule
    \multicolumn{2}{c|}{\textbf{Racetrack}} & \multicolumn{3}{c|}{{\textbf{$\boldsymbol{I_{T} [\downarrow] \text{ in } \si{\second}}$}}} & \multicolumn{3}{c|}{\textbf{$\boldsymbol{I_{C} [\downarrow] \text{ in } \si{\percent}}$}} & \multicolumn{3}{c}{\textbf{$\boldsymbol{I_{E} [\downarrow] \text{ in } \si{\per\km}}$}} \\
    & Name  & {APF} & \hspace{0.2cm}\underline{RaceMOP} &  {$\Delta_{\text{rel}}$ in \% } & {APF} & \hspace{0.2cm}\underline{RaceMOP} &  {$\Delta_{\text{rel}}$ in \% } & {APF} & \hspace{0.2cm}\underline{RaceMOP} &   {$\Delta_{\text{abs}}$}  \\
    \midrule
    \multirow{9}{*}{\rotatebox[origin=c]{90}{Training}} 
    &Budapest&57.14&51.62\textsuperscript{$\pm$0.45}&-9.67&22.47&0.4\textsuperscript{$\pm$0.36}&-98.22&0.0&0.0\textsuperscript{$\pm$0.0}&0.0\\%
    &Catalunya&58.86&53.67\textsuperscript{$\pm$0.61}&-8.81&24.05&0.81\textsuperscript{$\pm$0.35}&-96.62&0.0&0.06\textsuperscript{$\pm$0.12}&0.06\\%
    &Hockenheim&51.7&47.1\textsuperscript{$\pm$0.32}&-8.91&24.18&0.0\textsuperscript{$\pm$0.0}&100.0&0.0&0.0\textsuperscript{$\pm$0.0}&0.0\\%
    &Moscow&50.0&44.79\textsuperscript{$\pm$1.04}&-10.42&32.39&0.67\textsuperscript{$\pm$0.81}&97.93&0.0&0.0\textsuperscript{$\pm$0.0}&0.0\\%
    &N\"urburgring&63.4&57.88\textsuperscript{$\pm$0.37}&-8.71&24.66&0.13\textsuperscript{$\pm$0.29}&99.48&0.13&0.01\textsuperscript{$\pm$0.02}&-0.12\\%
    &Sakhir&62.19&57.08\textsuperscript{$\pm$0.31}&-8.22&12.5&0.53\textsuperscript{$\pm$1.18}&95.79&0.0&0.0\textsuperscript{$\pm$0.0}&0.0\\%
    &Sepang&68.2&62.81\textsuperscript{$\pm$0.64}&-7.9&23.6&0.0\textsuperscript{$\pm$0.0}&100.0&0.0&0.0\textsuperscript{$\pm$0.0}&0.0\\%
    &Spielberg&47.54&44.37\textsuperscript{$\pm$0.3}&-6.67&29.33&0.14\textsuperscript{$\pm$0.31}&99.52&0.0&0.03\textsuperscript{$\pm$0.03}&0.03\\%
    \midrule
    &All&57.38&52.41\textsuperscript{$\pm$0.46}&-8.65&23.55&0.33\textsuperscript{$\pm$0.19}&-98.61&0.02&0.01\textsuperscript{$\pm$0.01}&0.01\\%
\bottomrule
\end{tabular}
\caption{Performance metrics for \textit{training} racetracks with mean and standard deviation from five random seeds for RaceMOP.}
\label{tab:results_train_all}
\end{table*}

\begin{table*}[!t]
\centering
\begin{tabular}{P{0.1cm} l | S[table-format=2.2] S[table-format=2.2] S[table-format=2.2] | S[table-format=2.2] S[table-format=2.2] S[table-format=2.2] | S[table-format=2.2] S[table-format=2.2] S[table-format=2.2] }
\toprule
    \multicolumn{2}{c|}{\textbf{Racetrack}} & \multicolumn{3}{c|}{{\textbf{$\boldsymbol{I_{T} [\downarrow] \text{ in } \si{\second}}$}}} & \multicolumn{3}{c|}{\textbf{$\boldsymbol{I_{C} [\downarrow] \text{ in } \si{\percent}}$}} & \multicolumn{3}{c}{\textbf{$\boldsymbol{I_{E} [\downarrow] \text{ in } \si{\per\km}}$}} \\
    & Name  & {APF} & \hspace{0.2cm}\underline{RaceMOP} &  {Disparity} & {APF} & \hspace{0.2cm}\underline{RaceMOP} &  {Disparity} & {APF} & \hspace{0.2cm}\underline{RaceMOP} &   {Disparity}  \\
    \midrule
    \multirow{3}{*}{\rotatebox[origin=c]{90}{Test}} 
    &Brands Hatch&47.89&\bfseries45.38\textsuperscript{$\pm$0.36}&46.5&13.98&\bfseries0.28\textsuperscript{$\pm$0.38}&21.11&0.06&\bfseries0.0\textsuperscript{$\pm$0.0}&\bfseries0.0\\%
    &Melbourne&65.25&\bfseries60.64\textsuperscript{$\pm$0.26}&61.15&31.51&\bfseries0.28\textsuperscript{$\pm$0.38}&23.38&\bfseries0.0&0.01\textsuperscript{$\pm$0.02}&0.32\\%
    &Mexico City&52.52&47.69\textsuperscript{$\pm$0.53}&\bfseries47.55&35.62&\bfseries0.59\textsuperscript{$\pm$0.6}&29.87&\bfseries0.0&0.13\textsuperscript{$\pm$0.28}&0.29\\%
    &Sao Paulo&50.06&\bfseries45.87\textsuperscript{$\pm$0.25}&47.72&27.14&\bfseries0.55\textsuperscript{$\pm$0.58}&21.43&0.18&\bfseries0.01\textsuperscript{$\pm$0.02}&0.24\\%
    \midrule
    &All&53.93&\bfseries49.89\textsuperscript{$\pm$0.32}&50.73&26.21&\bfseries0.42\textsuperscript{$\pm$0.34}&23.68&0.05&\bfseries0.03\textsuperscript{$\pm$0.06}&0.21\\%
\bottomrule
\end{tabular}
\caption{Performance metrics for \textit{test} racetracks with mean and standard deviation from five random seeds for RaceMOP. The best results are bold.}\label{tab:results_test_all}
\end{table*}

This evaluation compares the performance of RaceMOP against the pure \gls*{apf} planner to demonstrate the effectiveness of the \gls*{rpl} approach. Our results are shown in Table \ref{tab:results_train_all} for the eight \textit{training} race tracks. First, it can be seen that RaceMOP improves the base policy w.r.t. all performance metrics:
\begin{itemize}
    \item Lab times are decreased by \SI{8.65}{\percent} to $I_T$\,=\,\SI{52.41}{\second} averaged for all training racetracks.
    \item Despite the faster lap times, RaceMOP manages to decrease unsuccessful overtaking substantially to \SI{0.33}{\percent} crashes per attempt, i.e., there is approx. one crash for 300 successful overtaking maneuvers.
    \item RaceMOP increases environment crashes $I_{E}$ for two tracks but reduces them when the \gls*{apf} struggled before.
\end{itemize}

Analysis of the lap times $I_C$ shows that the gains are similar on all training racetracks, ranging from \SI{-10.42}{\percent} for Moscow to \SI{-6.67}{\percent} for Spielberg.
For the \gls*{apf} planner, overtaking for Moscow, a racetrack with high average curvature, is most challenging with $I_C$\,=\,\SI{32.39}{\percent}.
RaceMOP is capable of reducing this number substantially by \SI{-97.93}{\percent} to \SI{0.67}{\percent}.
While overtaking, most collisions of RaceMOP happen at Catalunya with $I_C$\,=\,\SI{0.81}{\percent}. 
The best performance of RaceMOP is achieved for overtaking on the Hockenheim and Sepang racetracks where \textit{zero} crashes are recorded.

RaceMOP's performance is overall stable for different training runs as the standard deviations of $\pm\SI{0.46}{\percent}$ for $I_T$ and $\pm\SI{0.19}{\percent}$ for $I_C$ prove.
The higher deviations for $I_C$ on Sakhir and Moscow show that if there is some instability during training for these tracks, RaceMOP compensates by improved performance on all the other racetracks.
This may indicate that RaceMOP's representational capacity is limited by the used size of the \gls*{dnn} as performance gains on one track come at the cost of slightly more collision on another.

\subsection{Out-of-Distribution Benchmarking}
\label{subsec:results_2}
We evaluate the generalization capabilities of RaceMOP on four \textit{test} racetracks and in comparison to another approach: a disparity extender \cite{otterness2019disparity} with adaptive velocity planning to avoid collisions when there is no safe gap.
Our results in Table \ref{tab:results_test_all} align with the previously discussed improvements by RaceMOP.
While the disparity extender can race at higher velocities\footnote{Experiments have shown that the disparity extender as base policy does not work as well as the \gls*{apf} due to its rather aggressive and fast driving.} than the \gls*{apf} base planner, RaceMOP overcomes this disadvantage.
These results demonstrate RaceMOP's outstanding generalization capabilities, achieving an average ratio of $I_C$\,=\,\SI{0.42}{\percent}.
However, the environment crashes for Mexico City increase to $I_E$\,=\,\SI{0.13}{\percent}, which indicates that due to the lack of map data, there is a specific section that is not correctly interpreted by RaceMOP.

Table \ref{tab:additional_results} shows that in another out-of-distribution test with $v_k=0.85$, RaceMOP is still capable of robust overtaking as the crash rate increases marginally to $I_C$\,=\,\SI{0.86}{\percent}. 
These crashes occur now because RaceMOP sometimes overtakes when its velocity advantage is too small to finish the maneuvers before the opponent cuts the corner.
Lap times are decreased to $I_T$\,=\,\SI{51.98}{\second} as RaceMOP loses less time following the now faster-driving opponents before overtaking.

\subsection{Ablation Policy Fusion Method}

\begin{table}[!t]
\centering
\begin{tabular}{c | S[table-format=2.2]  S[table-format=2.2] S[table-format=2.2] }
    \toprule
   \textbf{Scenario} & {{\textbf{$\boldsymbol{I_{T} [\downarrow] \text{ in } \si{\second}}$}}} & {\textbf{$\boldsymbol{I_{C} [\downarrow] \text{ in } \si{\percent}}$}} & {\textbf{$\boldsymbol{I_{E} [\downarrow] \text{ in } \si{\per\km}}$}} \\
    \midrule
    $v_k=0.85$ & 51.98\textsuperscript{$\pm$0.48} & 0.86\textsuperscript{$\pm$0.62} & 0.01\textsuperscript{$\pm$0.01}\\
    $\mathtt{clip}(a_t)$ & 52.72\textsuperscript{$\pm$0.2} & 2.65\textsuperscript{$\pm$0.68} & 0.01\textsuperscript{$\pm$0.01}\\
    \bottomrule
\end{tabular}
\caption{Additional results for RaceMOP averaged for \textit{training} tracks.}
\label{tab:additional_results}
\end{table}

\label{subsec:results_3}
\begin{figure*}[htbp]
    \vspace{2mm}
    \centering
        \begin{subfigure}[b]{0.15\textwidth}
            \centering
            \includegraphics[trim={0mm 0mm 0mm 3mm},clip,height=7.3cm]{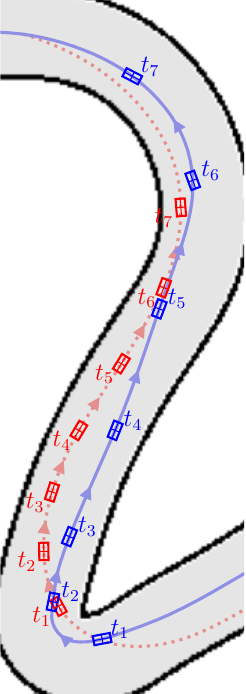}
            \caption{N\"urburgring.}
            \label{fig:ot_mos}
        \end{subfigure}
        \hfill
        \begin{subfigure}[b]{0.15\textwidth}
            \centering
            \includegraphics[trim={0mm 0mm 0mm 3mm},clip,height=7.3cm]{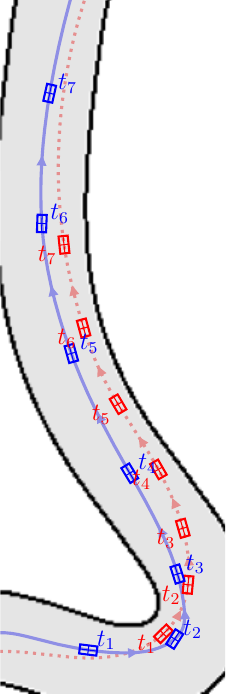}
            \caption{Sepang.}
            \label{fig:ot_sep}
        \end{subfigure}
        \hfill
        \begin{subfigure}[b]{0.15\textwidth}
            \centering
            \includegraphics[trim={0mm 5mm 0mm 3mm},clip,height=7.3cm]{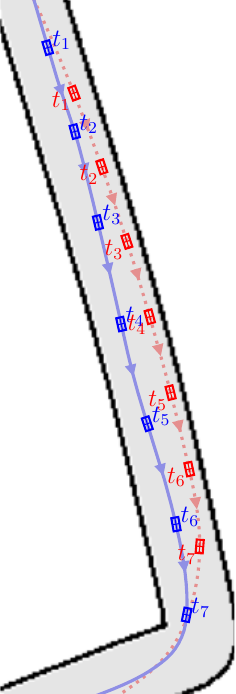}
            \caption{Melbourne.}
            \label{fig:ot_mel}
        \end{subfigure}
    \hfill
    \begin{minipage}[b]{0.43\textwidth}
        \centering
        \begin{subfigure}[b]{\textwidth}
                \centering
                \includegraphics[trim={2mm 3mm 9mm 0mm},clip, width=0.9\textwidth]{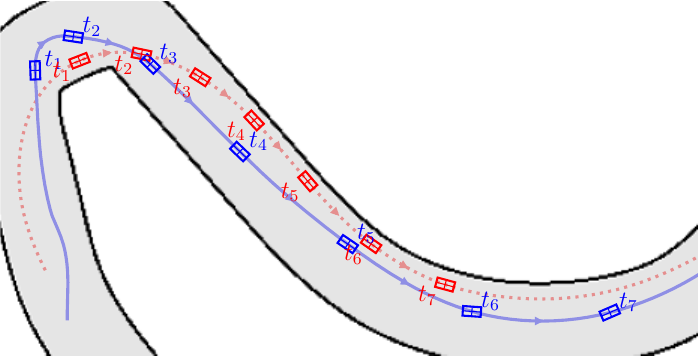}
                \caption{Sao Paulo.}
                \label{fig:ot_sao}
        \end{subfigure}
        \vfill
        \begin{subfigure}[b]{\textwidth}
            \centering
            \includegraphics[width=0.8\textwidth]{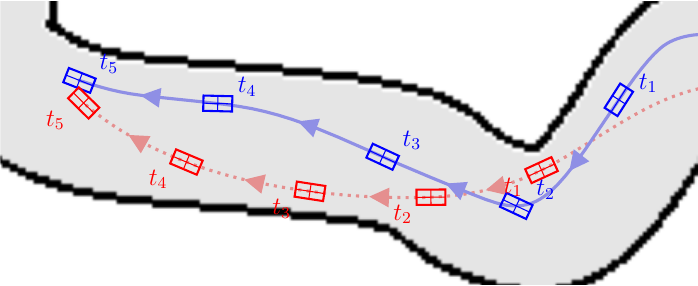}
            \caption{Moscow (crash).}
            \label{fig:crash_mos}
        \end{subfigure}
    \end{minipage}
    \caption{Exemplary overtaking maneuvers of RaceMOP for five different, replicated real-world racetracks where the ego vehicle (blue, full line) overtakes the opponent (red, dashed line), showing various strategic behaviors. Discrete timesteps $t_1,..., t_7$ of the vehicle's pose are given every 0.5\,$\si{\s}$.}
    \label{fig:composite}
\end{figure*}

We analyze the effect of the proposed truncated Gaussian distribution by replacing it with the clipped action sum as used in \cite{trumpp2023residual}.
The result in Table \ref{tab:additional_results} shows that this fusion method can achieve lap times close to RaceMOP. 
However, when comparing the crash risk, it is much increased to $I_C$\,=\,\SI{2.65}{\percent} from \SI{0.33}{\percent}.
This experiment demonstrates the substantial advantage of our proposed policy fusion method.

\subsection{Scenario Analysis}
We discuss RaceMOP's behavior with five examples.
Our analysis of various scenarios and racetracks shows that RaceMOP's advantage is a robust overtaking behavior at curves.
Typically, RaceMOP will approach a curve fast, break if the opponent switches sides when cutting the curve, and then accelerate quickly to pass the opponent.
This behavior is effective, affirming that RaceMOP is capable of long-horizon reasoning from local observations where other approaches typically require map data.

\subsubsection{Inside Overtaking (Fig.~\ref{fig:ot_mos} and Fig.~\ref{fig:ot_sep})}
The ego vehicle waits with the attempt as the opponent cuts the corner. After the apex, the overtaking becomes feasible as a gap opens at the inside, and the ego accelerates to overtake the opponent at a safe distance.

\subsubsection{High Velocity Overtaking (Fig.~\ref{fig:ot_mel})}
The ego uses its velocity advantage to overtake the opponent before a 90$^\circ$ curve by breaking late, requiring a careful estimation of its physical driving limit and the opponent's trajectory to finish the maneuver in time.

\subsubsection{Outside Overtaking (Fig.~\ref{fig:ot_sao})}
The opponent follows the wall in a curve closely. RaceMOP has learned to wait first and then takes advantage of this behavior, passing the opponent on the outside to finish the maneuver.

\subsubsection{Crash (Fig.~\ref{fig:crash_mos})}
This unsuccessful overtaking happens at a difficult section where RaceMOP first waits with the overtake. It then starts overtaking but incorrectly interprets the opponent's behavior, who cuts the corner. As the opponent closes in, the ego is not fast enough to finish the maneuver, leading to a crash.

\section{Limitations}\label{seq:limitations}
While RaceMOP shows strong abilities for generalization and can even overtake fast-driving opponents in out-of-distribution settings, we assume that the opponents do not show active defensive behavior nor that RaceMOP is surpassed.
While these assumptions align with the current literature, more various opponent behaviors should be evaluated in future work, e.g., multi-agent \gls{drl} with self-play.

For the time being, the difficult multi-agent setup prevents real-world experiments. First, a large racetrack with enough space for two cars side-by-side is required. Second, real-world training is impossible for the used 30e\textsuperscript{6} interactions. Thus, a direct sim-2-real deployment is required. RaceMOP's robust performance in simulation holds the promise that we can bridge the sim-2-real gap efficiently by using additional domain randomization and finetuning with real-world data. However, the current software stack does not incorporate such methods, preventing reliable real-world deployment.

%% file: source/conclusion.tex
\addtolength{\textheight}{-1cm}
\section{Conclusion}
We presented RaceMOP, a map-less online path planner for multi-agent autonomous racing.
RaceMOP combines an \gls*{apf} planner with a learned residual policy.
Our results demonstrate that RaceMOP clearly outperforms comparable, map-less planners with a collision ratio of $I_C$\,=\,\SI{0.33}{\percent} and $I_C$\,=\,\SI{0.42}{\percent} for training and test racetracks, respectively.
A key component of RaceMOP's strong performance is the presented novel method for policy fusion in probability space that uses a truncated Gaussian distribution.
RaceMOP's advantage is its ability to make long-horizon decisions despite lacking map data, where naive planners fail.

Our future work includes testing RaceMOP against various opponent strategies and eventually transferring our method to a real-world race with multiple F1TENTH cars. Additionally, we plan an ablation study on RaceMOP's architecture as incorporating an \gls*{rnn} can improve training efficiency and performance for interaction-rich environments \cite{trumpp2023efficient}. Another interesting aspect will be to incorporate continuous action-mapping techniques for safe learning \cite{theile2024learning} into our method.